\documentclass{article}

\usepackage[nonatbib,preprint]{neurips_2021}

\usepackage[utf8]{inputenc} 
\usepackage[T1]{fontenc}    
\usepackage{hyperref}       
\usepackage{url}            
\usepackage{booktabs}       
\usepackage{amsfonts}       
\usepackage{nicefrac}       
\usepackage{microtype}      
\usepackage{xcolor}         
\usepackage{amsmath}

\usepackage{graphicx}
\usepackage[ruled,noend]{algorithm2e}
\usepackage{float} 
\usepackage{caption}
\usepackage{csquotes}
\usepackage{subfig}
\usepackage{enumitem}
\usepackage[style=authoryear,maxcitenames=1,natbib]{biblatex}
\DeclareMathOperator*{\argmin}{argmin}

\title{Counterfactual Instances Explain Little}

\author{%
  Adam White, Artur d'Avila Garcez\\
  City Data Science Institute\\
  City, University of London, London, EC1V 0HB, UK\\
  \texttt{\{Adam.White, A.Garcez \}@city.ac.uk} \\
}

\begin{document}

\maketitle


\begin{abstract}
In many applications, it is important to be able to explain the decisions of machine learning systems. An increasingly popular approach has been to seek to provide \emph{counterfactual instance explanations}. These specify close possible worlds in which, contrary to the facts, a person receives their desired decision from the machine learning system. This paper will draw on literature from the philosophy of science to argue that a satisfactory explanation must consist of both counterfactual instances and a causal equation (or system of equations) that support the counterfactual instances. We will show that counterfactual instances by themselves explain little. We will further illustrate how explainable AI methods that provide both causal equations and counterfactual instances can successfully explain machine learning predictions.

\end{abstract}

\section{Introduction}
Machine learning systems are increasingly being used for automated decision making. It is important that these systems’ decisions can be trusted. This is particularly the case in mission critical situations such as medical diagnosis, airport security or high-value financial trading. But a machine learning system cannot be trusted simply on the basis of its accuracy with a test data set. The user also needs to be able to understand why the system is making its specific predictions. One possible solution is to treat machine learning systems as ‘black-boxes’ and to explain their input-output behaviour. Such approaches can be divided into two broad types: those providing global explanations of the entire system and those providing local explanations of single predictions. Local explanations are needed when a machine learning system’s decision boundary is too complex to allow for global explanations.

\citet{miller2018explanation} carried out a review of over 250 papers on explanation taken from the disciplines of philosophy, psychology and cognitive science. He states that perhaps his most important finding was that explanations are contrastive counterfactuals, seeking to answer the question ‘Why event E rather than event F?’ F is referred to as E’s foil and comes from a contrast class of events that were alternatives to E, but which did not happen. When a person asks for an explanation, the relevant contrast class is often not explicitly conveyed but instead is implicit in the explanatory question.

Multiple explainable AI (XAI) methods have been proposed that claim to provide ‘counterfactual explanations’ of single machine learning predictions. The explanation consists of either a single or multiple counterfactual instances. It is the argument of this paper that a satisfactory explanation must consist of both counterfactual instances and a causal equation (or system of equations) that supports the counterfactual instances. Counterfactual instances by themselves explain little.

The remainder of this paper is organised as follows: Section 2 provides an overview of counterfactual instances and highlights \citeauthor{karimi2021algorithmic}'s (\citeyear{karimi2021algorithmic}) argument that existing XAI methods often fail to successfully  generate actionable counterfactuals. Section 3 specifies the requirements for a satisfactory scientific explanation and argues that these are not satisfied by statements of counterfactual instances. Section 4 illustrates how XAI methods that provide both causal equations and counterfactual instances can satisfy these requirements. Section 5 discusses aspects of understandability, summarises the paper and discusses future directions of work.

\section{Counterfactual Instances}
A counterfactual instance specifies a close possible world in which, contrary to the facts, a person gets their desired outcome. For example, suppose that a banking machine learning system declined Mr Jones’ loan application and that the feature vector for Mr Jones was \{income: \$32,000, age: 45, education: graduate\}; then a counterfactual instance for Mr Jones might be \{income: \$35,000, age: 45, education: graduate\} where the \$3000 increase in salary is sufficient to flip Mr Jones to the desired side of the banking system’s decision boundary. \citet{wachter2017counterfactual} first proposed using counterfactual instances as explanations of single machine learning predictions. This type of explanation was intended primarily to be for the benefit of the data subject defined as ‘the natural person whose data is being collected or evaluated’. It has two main objectives (i) understanding: to help a person understand why a machine learning system made a particular prediction affecting them (ii) recourse: to specify what the person would need to change for the machine learning system to then produce the person’s desired outcome. Watcher et al. define a counterfactual explanation as having the form:

\begin{displayquote}{”Score $p$ was returned because variables $V$ had values ($v_1, v_2, \dots$) associated with them. If $V$ instead had values ($v'_1, v'_2, \dots$), and all other variables remained constant, score $p'$ would have been returned”}  \parencite[4]{wachter2017counterfactual} 
\end{displayquote}

\citeauthor{wachter2017counterfactual} state that their approach does not rely on ‘any knowledge of the causal structure of the world’. Nor is it intended to provide approximate explanations of a machine learning system’s algorithm or to provide feature scores. They note that their proposed type of explanation is different to those found in the previous ‘machine learning legal and ethics literature’ As we will illustrate, it is also substantially different to those proposed in the philosophy of science. They state that a similar type of explanation has been proposed by Nozick (1983, p.172:174)   within the philosophy of knowledge. Nozick is providing an analysis of the necessary conditions for propositional knowledge. He argues that satisfying the following subjunctive conditional may be necessary for a person $s$ to know proposition $p$:
\begin{displayquote}
If $p$ weren’t true, $s$ wouldn’t believe that $p$
\end{displayquote}
As Watcher et al. note, Nozick’s conditional can be understood as a counterfactual (given $p$ is true). Watcher et al.’s paper then goes directly from considering Nozick’s proposal to providing their definition of a counterfactual explanation. No philosophical argument is given to explain why Nozick’s conditional provides a justification for a form of explanation applicable to XAI.\\ 

There are now many XAI methods that attempt to generate ‘optimal’ counterfactual instances, for example \citet{karimi2020survey} review sixty counterfactual instance methods. The algorithms differ in the constraints they place and the attributes referenced in their loss functions (\citet{verma2020counterfactual}). Desiderata often include that a counterfactual instance is: (1) actionable – e.g. does not recommend that a person reduces their age (2) near to the original observation - common measures include Manhattan distance, L1 norm and L2 norm (3) sparse – only changing the values of a small number of features (4) plausible - e.g. the counterfactual instance must correspond to a high density part of the training data (5) efficient to compute. These XAI methods can be broadly viewed as having the following objective function for a counterfactual $x+\delta$ (this is adapted from \citet{karimi2021algorithmic} and \citet{ustun2019actionable}):
\[\delta \in   \argmin_\delta cost(\delta;x) \;\; s.t. \;  class(m(x^{CFE})) \neq class(m(x)) ,\; x^{CFE} = x + \delta \;, x^{CFE} \in P, \; \delta \in F \]

where $m$ is the machine learning system, $x$ is the observation, class$(m(x))$ is the label for prediction $m(x)$, $x^{CFE}$ is a nearest counterfactual, $F$ and $P$ are optional feasibility and plausibility constraints.

\citet{karimi2021algorithmic} show that counterfactual instances satisfying this function will often be suboptimal or infeasible; because they do not take account of the causal structure that determines the consequences of the person’s actions. The underlying problem is that unless all of the person’s features are causally independent of each other, then when the person acts to change the value of one feature, other downstream dependent features may also change. \citeauthor{karimi2021algorithmic}'s proposed solution uses Pearl’s interventionist calculus to reformulate the recourse problem. The objective function now being to find the minimum cost set of actions that will achieve the desired outcome, where these actions are modelled as interventions on the structural causal model that subsumes the relevant portion of the person’s world. In order to reliably compute a set of recourse actions, the structural causal model will need to be at least approximately true. Unfortunately, as \citeauthor{karimi2021algorithmic} note, the true causal model of the world is rarely known, and therefore their proposed solution currently has limited applicability. Nevertheless, the key point for this paper is that the counterfactual instances being produced by current XAI methods are often likely to fail the second of Watcher et al.’s objectives: identifying how a person should act in order to achieve their desired outcome. 

\section{Theories of Scientific Explanation}

The structure of a satisfactory scientific explanation has been extensively studied throughout the history of philosophy. The deductive-nomological (henceforth: D-N) account of scientific explanation once dominated the philosophy of science. According to this, to explain a phenomenon is to subsume it under general laws of nature. For systems with deterministic laws, successful scientific explanations were taken to have the form of the D-N model , as specified by Hempel and Oppenheim (1948):

\begin{figure}[!htbp]
  \includegraphics[width = 12cm]{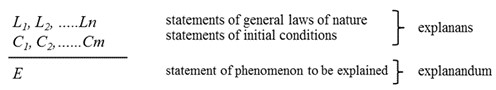}  
  \centering
  \caption{The schema for a DN-Explanation}
  \label{DN}
\end{figure} 

The explanans must include statements of general laws of nature that are essential to the derivation of the explanandum. The laws are general, in the sense that statements of the laws make no reference to particulars and are true without exception. The explanandum is explained by showing that it is an instantiation of these laws. A similar account of explanation was held to apply for systems with probabilistic laws, but it is sufficient for our purposes just to focus on D-N explanations. The D-N Model has been subject to some well-known counterexamples (Salmon, \citeyear{salmon2020scientific}, p.46-50). Consider, for example, the following deductive argument:

 \begin{displayquote}
Every man who regularly takes birth control pills avoids pregnancy\\
\underline{John Jones regularly takes birth control pills}\\
John Jones avoids becoming pregnant
 \end{displayquote}
 
This satisfies the criteria for being a D-N explanation but clearly fails to be explanatory. Such counterexamples have highlighted serious shortcomings with the D-N Model, including that: (i) irrelevant premises can be used to deduce and hence ‘explain’ an explanandum (ii) no temporal priority is required between the explanans and the explanandum (iii) perhaps most importantly, D-N explanations fail because a successful explanation of an event requires stating its causes.\\ \\
The dominant accounts of scientific explanation are now the counterfactual theories of causal explanation found in Woodward (\citeyear{Woodward}) and Halpern and Pearl (\citeyear{halpern2005causes}). This paper will primarily focus on the account found in Woodward’s book ‘Making Things Happen: A Theory of Causal Explanation’, which won the Lakatos Award for outstanding contribution to philosophy of science. Woodward and Pearl (\citeyear{Pearl}) share similar analyses, with Woodward  focussing more on the philosophy rather than the calculus of causality \footnote{See \citet{woodward2003critical} for an analysis of the differences between Woodward and Pearl's specification of an intervention}. However this papers arguments for XAI also apply, mutatis mutandis, to Pearl’s work.

Before providing Woodward’s definition of causal explanation, it is first necessary to briefly outline his notions of ideal intervention and invariance. For Woodward, causal relationships relate variables. Variables are properties or magnitudes that can have more than one value; and the values of variables are possessed by particular entities. (Woodward, 2003, p.39). 

An ‘ideal intervention’ on $X$ with respect to $Y$ exogenously changes the value of $X$, such that any change that occurs to the value of $Y$ occurs only because of the change in the value of $X$. Woodward’s specification of an ideal intervention involves an ‘intervention variable’ $I$ which acts like a ‘switch’. When $I$ is ‘switched on’:
\begin{enumerate} [topsep=0pt,itemsep=-1ex,partopsep=1ex,parsep=1ex]
\item $I$ causes $X$. $X$’s value is solely a function of $I$.
\item This means that all connections between $X$ and its pre-intervention causes are ‘broken’.
\item $I$ changes the value of $Y$, if at all, only by changing $X$.
\item $I$ does not alter the relationship between $Y$ and any of its causes $Z$ that are not on a directed path from $X$ to $Y$. (Woodward, \citeyear{woodward2008invariance}, p.202-203)
\end{enumerate}

\begin{figure}[!htbp]
  \includegraphics[width = 12cm]{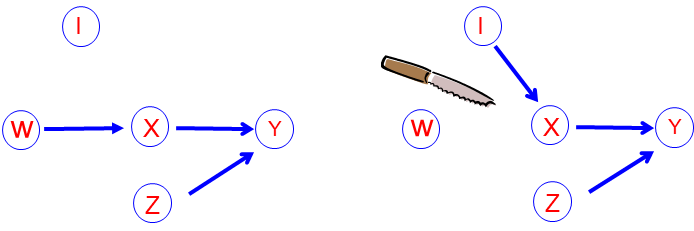}  
  \centering
  \caption{Example of an ideal intervention. $X$ and $Z$ are direct causes of $Y$. $I$ is an intervention variable for $X$ with respect to $Y$. When $I$ is switched on $X$ ceases to depend on $W$ and depends only on $I$.}
  \label{Ideal_intervention}
\end{figure}

An ideal intervention on $X$ with respect to $Y$ consists in $I$ being ‘switched on’ (see Figure \ref{Ideal_intervention}).
The relationship between $X$ and $Y$ is ‘invariant’ if it holds for at least one ‘testing intervention’. Let the relationship between $X$ and $Y$ be represented by the generalisation:\\
\[Y = G(X).\]
A testing intervention is an ideal intervention that changes the value of $X$ from, say, $x_0$ to $x_1$ and establishes that:
\[G(x_0) = y_0 \neq G(x_1) = y_1.\]
A necessary and sufficient condition for a generalisation between variables $X$ and $Y$ to represent a causal relationship is that it is invariant (Woodward, 2003, p.250). $X$ is a \emph{direct cause} of $Y$ with respect to a variable set $V$ if there is a possible change to the value of $X$ that will change the value of $Y$ when all other variables in $V$ (besides $X$) are held fixed. 

Woodward provides the following definition of a causal explanation for a simple system without multiple connections:
\begin{displayquote}
“Suppose that $M$ is an explanandum consisting in the statement that some variable $Y$ takes the particular value $y$. Then an explanans $E$ for $M$ will consist of (a) a generalization $G$ relating changes in the value(s) of a variable $X$ (where $X$ may itself be a vector or n-tuple of variables $X_i$ ) with changes in $Y$, and (b) a statement (of initial or boundary conditions) that the variable $X$ takes the particular value x. A necessary and sufficient condition for $E$ to be (minimally) explanatory with respect to $M$ is that (i) $E$ and $M$ be true or approximately so; (ii) according to $G$, $Y$ takes the value $y$ under an intervention in which $X$ takes the value $x$; (iii) there is some intervention that changes the value of $X$ from $x$ to where $x \neq x'$ , with $G$ correctly describing the value that $Y$ would assume under this intervention, where $y \neq y'$ ." (Woodward, 2003, p.203)
\end{displayquote}
Central to Woodward’s account is the requirement for an invariant generalisation. This is a causal equation in which the dependent variable is the effect and the independent variables are a complete set of its direct causes. This equation(s) support counterfactuals. An explanation reveals patterns of counterfactual dependence; and the purpose of counterfactual instances is to provide answers to a set of ‘what-if-things-had-been-done-differently’ questions. The generalisation(s) might only be true for a particular domain, and in such cases provides a local explanation. For Woodward, all causal claims are counterfactual and contrastive: ‘to causally explain an outcome is always to explain why it, rather than some alternative, occurred’(Woodward, 2003, p.146).

This paper endorses Woodward’s theory as specifying the structure of a satisfactory XAI explanation. A machine learning system’s prediction $y$ is caused by the values of the features of observation $x$. A satisfactory explanation therefore needs to specify (i) an approximately true invariant generalisation relating the prediction to the input features and (ii) a set of counterfactual instances that are supported by the invariant generalisation. The resulting explanation reveals the (approximate) direct causal relationships that the machine learning system has learnt between the features of $x$ and prediction $y$.

It can be argued that counterfactual instances are a type of contrastive causal explanation (e.g. see \citet{karimi2020survey}). They identify a subset of the causes of a machine learning system $m$ predicting class $l$ for observation $x$, and specify counterfactual changes in the values of those causes that would result in $m$ predicting class $l'$. But these are, at best, shallow causal explanations. They have multiple failings. For example, they do not identify: the relative importance of different causes, each cause's functional form or any interactions between the causes.  Furthermore, they might not state all the key causes of an event, Consider again Mr Jones’ failed loan application. It may have been the case that the machine learning system was completely insensitive to any changes in Mr Jones' salary until it exactly reached \$35,000, or perhaps Mr Jones only needs to increase his salary to \$34,000 if he also enrols on an MBA. Providing only counterfactual instances is clearly insufficient to understanding the machine learning system's local behaviour. Imagine another science, say physics, treating a statement of counterfactual instances as an explanation, rather than seeking to discover the governing equation(s).


\section{Current XAI Methods}

\begin{figure}[!tbp]
  \includegraphics[width = 12cm]{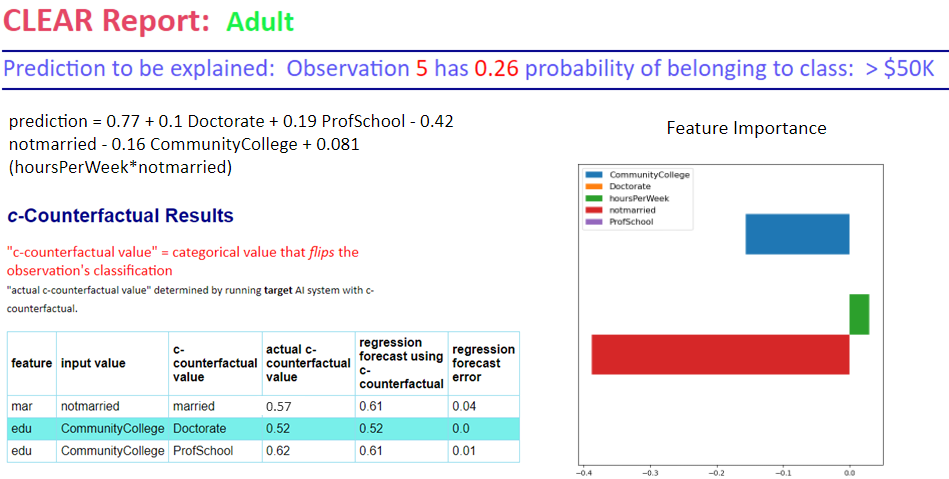}  
  \centering
  \caption{Extracts from a CLEAR \textit{Image} report. This explains a prediction $y$ made by a neural network for an observation $x$ from the Adult dataset. CLEAR states the causal equation that applies to the neighbourhood around $x$. Three counterfactuals are identified; for example intervening to change $x$ to being 'married', changes $y$ to 0.57. CLEAR's causal equation estimates that $y$ will change to 0.61, hence there is a fidelity error of 0.04.}
  \label{CLEAR_report}
\end{figure}

Two XAI methods will now be outlined, in order to illustrate how satisfactory causal explanations can be produced.\footnote{It is unclear as to the extent to which other prominent XAI methods could be adapted to satisfy Woodward’s specification e.g. LORE (\cite{guidotti2018local})  generates local decision trees rather than equations; Kernel SHAP's (\cite{lundberg2017unified}) regression coefficients are SHAP values and therefore do not directly calculate the effect of changing a feature's value on the machine learning system's prediction.} . The first is LIME: \textbf{L}ocal \textbf{I}nterpretable \textbf{M}odel-agnostic \textbf{E}xplanations (\cite{LIME}), which generates a causal equation, however LIME would have to be enhanced to produce counterfactual instances. The second is CLEAR: \textbf{C}ounterfactual\textbf{ L}ocal \textbf{E}xplanations vi\textbf{A} \textbf{R}egression (\cite{white2019measurable,white2021contrastive}) which was explicitly developed to satisfy Woodward’s definition. Both explain single predictions by performing local weighted regressions in which the dependent variable is the prediction $y$ and the independent variables are the input features $x$, a subset of which are the direct causes of $y$. The regressions are carried out on a synthetic dataset, with the values of each synthetic observation $x'$ being generated by replacing subsets of the observation x with random samples from parametric distributions and passing $x'$ through the machine learning system to determine the corresponding $y'$. Hence both methods are assuming that the values of each direct cause can be independently changed (a requirement of Woodward's theory). CLEAR was developed after LIME, with key differences including that: CLEAR discovers actual counterfactual instances by sampling $m$, includes the counterfactual instances in its regression dataset, and uses GAN generated images for explaining image classifications. CLEAR also measures its fidelity i.e. how faithfully it mimics the input-out behaviour of the machine learning system it seeks to explain. Figure \ref{CLEAR_report} shows extracts from a CLEAR explanation for the classification probability of an observation taken from the Adult dataset. This illustrates how an XAI method can satisfy Woodward's requirements, with CLEAR providing a causal equation, counterfactual instances and reporting fidelity statistics (see \citet{white2019measurable} for a detailed comparison of LIME and CLEAR; see \citet{white2021contrastive} for how CLEAR uses a GAN to provide contrastive explanations of image classifications). 

\section{Discussion}

 To explain an event is “to provide information about the factors on which it depends and exhibit how it depends on those factors” (Woodward, \citeyear{Woodward}, p.204). This requires counterfactual explanations that reveal the causal structure producing the event. This paper has argued that counterfactual instances by themselves do not do this. They need to be supported by either a causal equation or system of equations.
 
 This paper’s focus is in the opposite direction to \citeauthor{karimi2021algorithmic}. Their focus is on recourse, whilst this paper’s focus is on providing an understanding of why the machine learning system made its prediction. \citeauthor{karimi2021algorithmic}’s criticism is that XAI methods do not provide actionable counterfactual instances, as they do not take account of the causal structure that generates the machine learning system's input data. This paper's criticism concerns counterfactual instances being insufficient to explain the direct causal structure linking the values of an observation $x$'s features with the machine learning system's prediction. Existing XAI counterfactual instance methods can be relevant here, as they help to illustrate the direct causal relationships that the machine learning system has learnt; but they need a supporting causal equation.
 
 A concern with providing Woodward’s form of explanation is understandability. In order to satisfy the requirement that the causal equation(s) is at least approximately true, the equation(s) may need to be too complex for non-technical users to understand. For example, CLEAR's equations may include logistic functions, interaction terms, exponents and so forth. The solution to this is to recognise that different types of users require different levels of explanation. A data scientist may want to see the causal equation that most faithfully mimics the local behaviour of the machine learning system. Sometimes these equations may include terms with functions that contradicts their understanding of the domain, suggesting that the machine learning system’s prediction is untrustworthy. By contrast a bank customer may only want to know which features most impacted the machine learning system’s prediction. The issue then becomes one of deciding how much of the full explanation to provide to a particular user, in what format, and when to \emph{drill down} into progressively more complex explanations. For example, for the bank customer, perhaps only a bar chart showing the relative weightings of the most important features plus some counterfactual instances might be sufficient.

 Since \citeauthor{wachter2017counterfactual}'s paper, considerable resources have been devoted to developing counterfactual instance methods. Greater focus is now needed on developing methods for discovering the direct causal relations that generate these counterfactual instances, and for making them useful in practice as an explanatory tool that can help increase trust in machine learning systems.

\clearpage

\printbibliography

\end{document}